# Developing a cardiovascular disease risk factor annotated corpus of Chinese electronic medical records

Jia Su[1], Bin He[2], Yi Guan[3,*], Jingchi Jiang[4], Jinfeng Yang[5]

Address: [1, 2, 3, 4] Language Technology Research Center, School of Computer Science and Technology, Harbin Institute of Technology, Harbin, China; [5] School of Software, Harbin University of Science and Technology, Harbin, China

Email: [1] sujiahit@gmail.com; [2] hebin_hit@foxmail.com; [3] guanyi@hit.edu.cn; [4] jiangjingchi0118@163.com; [5] fondofbeyond@163.com

[*] Corresponding author: Yi Guan, Room 803, Zonghe Building, Harbin Institute of Technology, No. 92 West Dazhi Street, Harbin, Heilongjiang, China; guanyi@hit.edu.cn; +86-18686748550

## ABSTRACT

**Background:** Cardiovascular disease (CVD) has become the leading cause of death in China, and most of the cases can be prevented by controlling risk factors. The goal of this study was to build a corpus of CVD risk factor annotations based on Chinese electronic medical records (CEMRs). This corpus is intended to be used to develop a risk factor information extraction system that, in turn, can be applied as a foundation for the further study of the progress of risk factors and CVD.

**Results:** We designed a light annotation task to capture CVD risk factors with indicators, temporal attributes and assertions that were explicitly or implicitly displayed in the records. The task included: 1) preparing data; 2) creating guidelines for capturing annotations (these were created with the help of clinicians); 3) proposing an annotation method including building xthe guidelines draft, training the annotators and updating the guidelines, and corpus construction. Meanwhile, we proposed some creative annotation guidelines: (1) the under-threshold medical examination values were annotated for our purpose of studying the progress of risk factors and CVD; (2) possible and negative risk factors were concerned for the same reason, and we created assertions for annotations; (3) we added four temporal attributes to CVD risk factors in CEMRs for constructing long term variations. Then, a risk factor annotated corpus based on de-identified discharge summaries and progress notes from 600 patients was developed. Built with the help of clinicians, this corpus has an inter-annotator agreement (IAA) $F_1$-measure of 0.968, indicating a high reliability.

**Conclusion:** To the best of our knowledge, this is the first annotated corpus concerning CVD risk factors in CEMRs and the guidelines for capturing CVD risk factor annotations from CEMRs were proposed. The obtained document-level annotations can be applied in future studies to monitor risk factors and CVD over the long term.

## BACKGROUND

**Introduction**

Cardiovascular disease (CVD) has become the first death cause throughout the world; there were approximately 17.5 million deaths from CVD in 2012, most of which occurred in low- and middle-income countries [1]. In China, CVD occupies the leading position among causes of death and is responsible for 2 out of every 5 deaths [2]. This situation deeply affects the health of the Chinese people and is a heavy burden on society. Fortunately, most CVD can be prevented by controlling the malleable risk factors such as specific medical conditions and the adoption of unhealthy life-styles at early stages [3]. A risk factor is a pattern of behavior or physical characteristic of a group of individuals that increases the probability of the future occurrence of one or more diseases in that group relative to comparable groups without or with different levels of the behavior or characteristic [4]. Risk factors, including specific medical conditions such as hypertension and hyperglycemia/diabetes, unhealthy life-style choices such as smoking and alcohol abuse, and other factors such as age and family history, can have prominent effects on the progress of CVD [3, 5]. Therefore, monitoring these risk factors constitutes an important approach in avoiding CVD.

A Chinese electronic medical record (CEMR) is a storage medium suitable for extracting CVD risk factors and monitoring. Actually, an electronic medical record (EMR) stores all health care data and information in electronic formats, along with the associated information processing and knowledge support tools necessary for the managing the health enterprise system [6]. The availability of large amounts of individual health narratives in CEMRs make this resource suitable for study by natural language processing (NLP) techniques especially information extraction (IE) techniques [7]. In 2010, the ministry of health in China published the basic norms of medical records writing [8] and the basic norms of electronic medical records [9], making these data more normative. Together, these characteristics make CEMR an effective medium for studies that involve extracting CVD risk factors. Some related works [7, 10—21] have been performed, but no studies have been conducted on CVD risk factors based on CEMRs. To do this, we designed a task to extract CVD risk factors from CEMRs.

To perform the extraction, we developed a CVD risk factor annotated corpus based on CEMRs because in the biomedical field the utilized corpora are far less than other open fields and a specific corpus is critically important in building an IE system. The purpose of this corpus is to act as the basis for developing an automatic risk factor extraction system. Subsequently, a monitoring platform could be established based on this extraction system that can help supervise CVD risk factors over time. Furthermore, based on the risk factors (along with other health information) that are comprehensively stored over long durations, a method that could predict the trend of each risk factor, help manage chronic diseases (such as hypertension and diabetes) and estimate the progress of CVD could also be included in the platform. To build the corpus, we proposed a light annotation task [22]. As the first step, we annotated 600 patients' de-identified discharge summaries and progress notes from CEMRs.

Our work is similar to the 2014 i2b2/UTHealth risk factor annotation shared task [23]. We adopted some technologies from that task, but also made some creations: (1) we proposed 12 CVD risk factors to be annotated including alcohol abuse, chronic kidney disease which were not considered in that task based on the advices of medical experts after considering the characteristics of CEMRs; (2) positive, possible, negative information and under-threshold examination values form a part of one's health condition and can be used to develop a long term supervision system, so these information were appended for our annotations; (3) for a long term monitoring, time information is critical, so we created temporal attributes for marking the occurrence time of risk factors in CEMRs.

**Related work**

*Related works based on English EMRs*

The 2006 Informatics for Integrating Biology and the Bedside (i2b2) shared task focused on identifying patients' smoking statuses from medical discharge records. In this task, 928 records covering five categories were annotated by two pulmonologists [24]. The 2008 i2b2 Obesity Challenge was an organized competition intend to find ways to recognize obesity and comorbidities from discharge summaries and classify them into four classes: Present, Absent, Questionable, and Unmentioned. An annotated data set was provided [25]. In 2009, the challenge focused on extracting medication information from medical records, including the names of medications, their dosages, modes and frequencies of administration, treatment durations, and reasons for administration [26, 27]. The challenge included a set of annotated discharge summaries. In 2010, the challenge involved a concepts, assertions, and relations identification task in which participants were given an annotated gold-standard corpus for system training [28]. The Sixth i2b2 Natural Language Processing Challenge concerned the issues involved in recognizing temporal relations in clinical records; it provided a corpus of annotated discharge summaries with temporal information [29, 30]. Subsequently, the 2014 i2b2/UTHealth NLP project focused on identifying risk factors for Cardiac Artery Disease in the narrative texts of EMRs, providing a set of 1,304 annotated medical records [31, 32]. In 2016, the challenge was to classify psychotic patients into four severities based on their neuropsychiatric clinical records; 433 annotated records were provided for training [33].

Projects such as the ShARe/CLEF eHealth Evaluation Lab 2013, were devoted to solving the difficulties involved in understanding the professional expressions (such as non-standard abbreviations, and ward-specific idioms) that clinicians use when describing their patients [34, 35]. This project provided annotated corpora for system building. Another important project was undertaken during SemEval 2015. Its clinical TempEval sub-task was similar to the i2b2 2012 NLP shared task in that participants were asked to find ways to recognize temporal information, clinical events, and their relations in clinical narratives. This project used a manually annotated corpus based on 600 clinical notes and pathology reports [36]. Another SemEval 2015 subtask involved analyzing clinical text, which involves named entity recognition and template field completion. This subtask used the ShARe corpus of annotated clinical text [35, 37]. Meystre et al. [38] proposed a new IE system for a congestive heart failure performance measure based on clinical notes from 1,083 Veterans Health Administration patients. Domain experts' annotated notes were created to act as a gold standard. Ford et al. [39] reported that information extracted from text in EMRs does improve case detection when combined with proper coding.

*Related works based on CEMRs*

Wang et al. [16] focused on recognizing and normalizing the names of symptoms in traditional Chinese medicine EMRs. To perform judgements, this system used a set of manually annotated clinical symptom names. Jiang et al. [14] proposed a complete annotation scheme for building a corpus of word segmentation and part-of-speech (POS) from CEMRs. Yang et al. [11] focused on designing an annotation scheme and constructing a corpus of named entities and entity relationships from CEMRs; they formulated an annotation specification and built a corpus based on 992 medical discharge summaries and progress notes. Lei [17] and Lei et al. [18] focused on recognizing named entities in Chinese medical discharge summaries. They classified the entities into four categories: clinical problems, procedures, labs, and medications. Finally, they annotated an entities corpus based on CEMRs. Xu et al. [19] studied a joint model that performed segmentation and named entity recognition in Chinese discharge summaries and built a set of 336 annotated Chinese discharge summaries. Wang et al. [20] researched the extraction of tumor-related information from Chinese-language operation notes of patients with hepatic carcinomas, and annotated a corpus contains 961 entities. He et al. [21] proposed a comprehensive corpus of syntactic and semantic annotations from Chinese clinical texts.

Despite the similar intent of these works, research into extracting CVD risk factors from CEMRs has not yet been studied. Meanwhile, for the IE tasks in the biomedical field, the number of accessible corpora is far fewer than those for more general extractions. However, corpora are important for building IE system. Thus, constructing a CVD risk factor annotated corpus is both a necessary and fundamental task. Moreover, unlike annotation tasks for texts that require less specialized knowledge, linguists require the assistance of medical experts to perform annotations in the biomedical field.

**MATERIALS AND METHODS**

**A light annotation task**

Compared with traditional NLP tasks such as segmentation, POS tagging, parsing, and semantic analysis, annotating CVD risk factors from CEMRs is a task that is both distinctive and light. As Stubbs says [22], we need only create a light annotation task for risk factor annotation rather than implementing all the NLP tasks. Therefore, based on the annotation trials conducted by Stubbs and Uzuner [32], we built a light annotation task that focuses solely on annotations of CVD risk factors with indicators, temporal attributes, and assertions and that does not require other NLP tasks. Meanwhile, an exhaustive annotation strategy— (we annotated all the occurrences of a CVD risk factor in CEMR narratives no matter how many times they appeared) —was applied. Notably, during the annotation trials, the increased time consumption caused by the exhaustive annotation strategy was subsequently offset by a higher level of inter-annotator agreement (IAA) and reduced difficulty for the annotators.

**Data**

We obtained a snapshot of medical records from the Second Affiliated Hospital of Harbin Medical University (a large general hospital that offers clinical services, medical education, scientific research, disease prevention, healthcare and rehabilitation) for all of 2012. The data included images of the medical records for approximately 140,000 patients from 35 departments and 87 sub-departments, ranging from pediatrics to the (Intensive Care Unit) ICU. To function as annotation tasks for CVD risk factors, we selected a subset of CEMRs from 600 patients composed of 344 randomly selected cardiovascular patients, 190 cardiovascular surgery patients, and 66 other departments. Each patient's medical records contained a series of documents consisting of their discharge summary, progress notes, medical examination reports, electrocardiograms etc. The discharge summaries and progress notes were regarded as the most important free text in these records [7]. A discharge summary is used to summarize the entire therapeutic process and treatment outcome, while progress notes record the clinical manifestations, medical examinations and treatment periodicity. Therefore, we regarded discharge summaries and progress notes for the 600 patients described above as suitable for annotation.

Next, the records were preprocessed as follows: (1) we used an optical character recognition (OCR) tool, "Tesseract," [40] to convert the original record images into text; (2) we manually fixed errors after the OCR process was complete and removed identifying information such as patient names, addresses, hospital IDs and doctor names; (3) we encoded the text into Extensible Markup Language (XML) format and added a title section using an XML node. Figure 1 shows an example of a progress note in XML format after preprocessing and ready for annotation.

```xml
<?xml version="1.0" encoding="UTF-8"?>
<progress>
        首次病程记录
    <主诉>    → chief complaints
        女，55岁，主因"发作性胸闷气短4年，加重20天"，门诊以"冠心病 不稳定型心绞痛"收入院。
    </主诉>
    <病例特点>    → case characteristics
        1、55岁女患，否认高血压、糖尿病、肝炎结核病史，否认吸烟饮酒史。阑尾炎手术史。
        2、患者入院前4年无诱因出现胸闷气短症状，无心前区及左肩部放射性疼痛和咽部紧缩感，伴出汗、乏力，无恶心、呕吐、头晕、咳嗽、咳痰。症状持续约10余分钟，休息后逐渐自行缓解。此后上述症状偶有发作，病程中服用速效救心丸、稳心颗粒、拜阿司匹灵（具体用量不详）。入院前20天，患者爬楼梯后出现上述症状，性质较前剧烈，持续时间延长，自服硝酸甘油1片后逐渐缓解。为求明确诊治于今日来我院就诊。患者患病以来，饮食二便正常，睡眠良好。
        3、查体：脉搏：73次/分，呼吸：18次/分，Bp：135/90mmHg，步入病室，发育正常，神清语明，自主体位，查体合作。眼睑无浮肿，口唇无发绀，颈静脉无怒张。双肺呼吸音清，未闻及干湿啰音。HR:73次/分，心律齐，未闻及早搏，各瓣膜区未闻及杂音。腹平软，无压痛及反跳痛，肝脾未及。双下肢无浮肿。
        4、辅助检查：ECG示：窦性心律，Ⅲ、aVR及V1v4导联ST段水平型下移，T波倒置
    </病例特点>
    <临床初步诊断>    → preliminary clinical diagnosis
        冠心病
        不稳定型心绞痛
    </临床初步诊断>
    <诊断依据>    → diagnosis basis
        1、55岁女患，发作性胸闷气短病史；
        2、否认高血压、糖尿病、肝炎结核病史，否认吸烟饮酒史；
        3、查体：Bp：135/90mmHg 双肺呼吸音清，HR：73次/分，心律齐，未闻及早搏及杂音，腹平软，无压痛及反跳痛，肝脾未及，双下肢无浮肿；
        4、心电图：窦性心律，Ⅲ、aVR及V1V4导联ST段水平型下移，T波倒置。
    </诊断依据>
    <鉴别诊断>    → differential diagnosis
        1、冠心病心肌梗死：疼痛性质剧烈，时程长，硝酸甘油不易缓解，心肌酶学和心电图有助鉴别
        2、主动脉夹层：患者多有常年高血压病史，发病时疼痛呈撕裂样，持续不缓解，超声及主动脉CTA有助鉴别。
    </鉴别诊断>
    <诊疗计划>    → assessment and planning
        1、抗凝，抗血小板；
        2、扩张冠脉，改善心肌供血；
        3、调节血脂，稳定冠脉粥样斑块；
        4、完善辅助检查：血、尿常规、凝血像、生化系列、肝炎系列、胸部正侧位片、心脏超声、holter；
        6、建议择期行 cAG+PCI 以进一步明确诊治。
    </诊疗计划>
</progress>
```

**Figure 1.** A sample progress note after preprocessing

**Annotation guidelines**

A light annotation task involves annotating the CVD risk factors with indicators, temporal attributes and assertions from the narratives in the CEMRs. Based on the CEMR characteristics and clinician suggestions, the guidelines for annotating this information are presented as follows.

*CVD risk factors and indicators*

An indicator is used to indicate the existence of a risk factor that may not be explicitly recorded in the narratives of CEMRs but exist in a cryptic form (e.g., "最高血压达 150/100 mmHg" (the highest blood pressure (Bp) is 150/100 mmHg) indicates a hypertensive patient). Explicitly mentioned risk factors and indirect expressions such as tests or treatments that can indicate the existence of risk factors are given equal status. Even indirect information (e.g., quantitative values from medical examinations) can be meaningful because it captures additional details about a patient's condition. With the assistance of medical experts, we proposed to annotate a set of CVD risk factors that include Overweight/Obesity (O2), Hypertension, Diabetes, Dyslipidemia, Chronic Kidney Disease (CKD), Atherosis, Obstructive Sleep Apnea Syndrome (OSAS), Smoking, Alcohol Abuse (A2), Family History of CVD (FHCVD), Age and Gender, and exploited the indicators of these risk factors. Table 1 lists all 12 types of risk factors, and this is the first time concerning those risk factors in CEMRs. The risk factors are in the left column and the indicators are on the right.

**Table 1** CVD risk factors and their indicators

| Risk factors | Indicators |
| --- | --- |
| O2 | • **Mention**: A diagnosis of patient overweight or obesity, e.g., "肥胖" (obesity) |
| Hypertension | • **Mention**: A diagnosis or history of hypertension, e.g., "高血压病史 20 年" (a history of hypertension for 20 years) |
| | • **High Bp**: A measurement of Bp or a description of the |
|  | patient's high Bp, e.g., Bp 150/100 mmHg |
|  | • **Regulate Bp**: A description of Bp regulation or unsuccessful regulation, e.g., "控制血压" (regulating Bp) |
|  | • **Blood pressure drug**: Patient takes medicine which is confirmed to control Bp, e.g., "口服降压药" (patient is taking hypotensor) |
| Diabetes | • **Mention**: A diagnosis or a history of diabetes, e.g., "糖尿病" (diabetes) |
|  | • **High blood glucose**: A measurement of blood glucose or a description of the patient's high blood glucose, e.g., blood glucose: "随机血糖：14.5 mmol/L" (Random Blood Glucose (RBG): 14.5 mmol/L) |
|  | • **Regulate blood glucose**: A description of blood glucose regulation or unsuccessful regulation, e.g., "调节血糖" (regulating the glucose) |
|  | • **Hypoglycemic drug**: Patient takes medications confirmed to control blood glucose, e.g., "规律用胰岛素" (patient takes insulin regularly) |
| Dyslipidemia | • **Mention**: A diagnosis of dyslipidemia, hyperlipidemia or a history of hyperlipidemia, e.g., "高血脂史" (a history of hyperlipidemia) |
|  | • **High blood lipids**: A measurement of blood lipids or a description of the patient's high blood lipids, e.g., TG (triglyceride): 1.96 mmol/L |
|  | • **Regulate blood lipids**: A description of blood lipids regulation or unsuccessful regulation, e.g., "降脂" (to lower serum lipids) |
|  | • **Lipid-lowering drug**: Patient takes medicine that is confirmed to control blood lipids, e.g., "治疗计划：立普妥 20 mg Qn po" (Treatment plan: Lipitor (atorvastatin calcium) 20 |
| | mg, take orally, once per night) |
| CKD | • **Mention**: A diagnosis of CKD, e.g., "慢性肾炎" (chronic nephritis) |
| Atherosis | • **Mention**: A diagnosis of atherosclerosis or atherosclerotic plaque, e.g., "冠脉粥样斑块" (atherosclerotic plaque) |
| OSAS | • **Mention**: A diagnosis of OSAS, e.g., "阻塞型睡眠呼吸暂停综合症" (OSAS) |
| Smoking | • **Mention**: Smoking or a patient history of smoking, e.g., "吸烟 40 余年" (smoking over 40 years) <br> • **Smoking cessation**: A description of smoking cessation or that the patient has not yet quit smoking, e.g., "未戒烟" (the patient has not yet quit smoking) <br> • **Smoking amount**: A description of how much the patient smokes, e.g., "每天 20 支" (20 cigarettes per day) |
| A2 | • **Mention**: Alcohol abuse, e.g., "长期大量饮酒史" (a long history of heavy drinking) <br> • **Drinking amount**: A description of how much the patient drinks, e.g., "2 两/日"（数值要大于 1）(100 grams of white spirit consumption per day (the number should be over 50 g)) |
| FHCVD | • **Mention**: Patient has a family history of CVD or has a first-degree relative (parents, siblings, or children) who has a history of CVD, e.g., "哥哥有冠心病病史" (the patient's brother has a history of CVD) |
| Age | • **Mention**: The age of the patient, e.g., "66 岁" (66 years old) <br> • **Age group**: The age group of the patient, e.g., "老年"(elderly) |
| Gender | • **Mention**: The gender of patient, e.g., "女性" (female) |

Notably, with the goal of being able to construct a timeline of CVD risk factors, we annotated all the quantitative values from medical examinations regardless of whether they exceeded the threshold (e.g., a patient whose Bp is 120/80 mmHg is also annotated for hypertension, even though the measurement is below the 140/90 mmHg criterion [5]). This was done so that all the test values would be extracted and we could later build a clear picture of changes in risk factors over time.

*Temporal attributes*

To construct a health condition timeline, collecting temporal annotations is essential. Considering the time at which the indicators occurred and the characteristics of CEMRs, we proposed to divide the risk factors into four time-dependent categories: before the duration of hospital stay (DHS) (the risk factor occurred before the DHS), during the DHS, after the DHS, and continuing (the risk factor is continuous). For instance, "平时血压 130/90 mmHg" (a patient whose ordinary Bp is 130/90 mmHg) is regarded as the "High Bp" indicator of hypertension with time before the DHS; "查体：Bp 130/80 mmHg" (physical examination: Bp 130/80 mmHg) indicates that the "High Bp" of Hypertension was made during the DHS; "出院医嘱：调节血糖" (doctor advice to a patient after discharge: to regulate blood glucose) indicates that the diabetes indicator of "Regulate Blood Glucose" occurred after the DHS; and "身材肥胖" (obesity) which is usually unchangeable over the short term would be annotated as a "Mention" of O2 with time continuing. In this way, changes in risk factors can be clearly presented. For example, "no indicators of diabetes were presented during the previous DHS, but evidence shows that the patient exhibited diabetic indicators before the next DHS; therefore, the diabetes occurred between the two DHSs." Notably, age and gender were not included in the temporal annotations.

*Assertions*

In contrast to the works of Stubbs and Uzuner [32], we are the first to propose assertions of risk factors. For example, "无糖尿病病史" (patient does not have a history of diabetes) needs to be considered, because such text show that the patient did not previously have diabetes. Based on whether the assertion of the risk factor actually occurred on the patient, we created two modifiers: associated or not associated with the patient. Further, risk factors associated with the patient are divided into three categories: present, absent and possible. Overall, assertions can be summarized as follows:

- Present: the risk factor definitely occurred on the patient, e.g., "平时血压 130/90 mmHg" (the patient's ordinary Bp is 130/90 mmHg)

- Absent: the risk factor was considered for the patient, but was negative, e.g., "无糖尿病病史" (the patient does not have a history of diabetes)

- Possible: the risk factor may possibly have occurred on the patient, e.g., "临床初步诊断：糖尿病" (Primary diagnosis: diabetes)

- Not associated with the patient: the risk factor occurred on others, e.g., "弟患糖尿病" (the patient's brother has diabetes)

**Annotation method**

The annotation method involves three major tasks: drafting the guidelines, training the annotators and updating the guidelines, and corpus construction. These tasks can be seen in Figure 2.

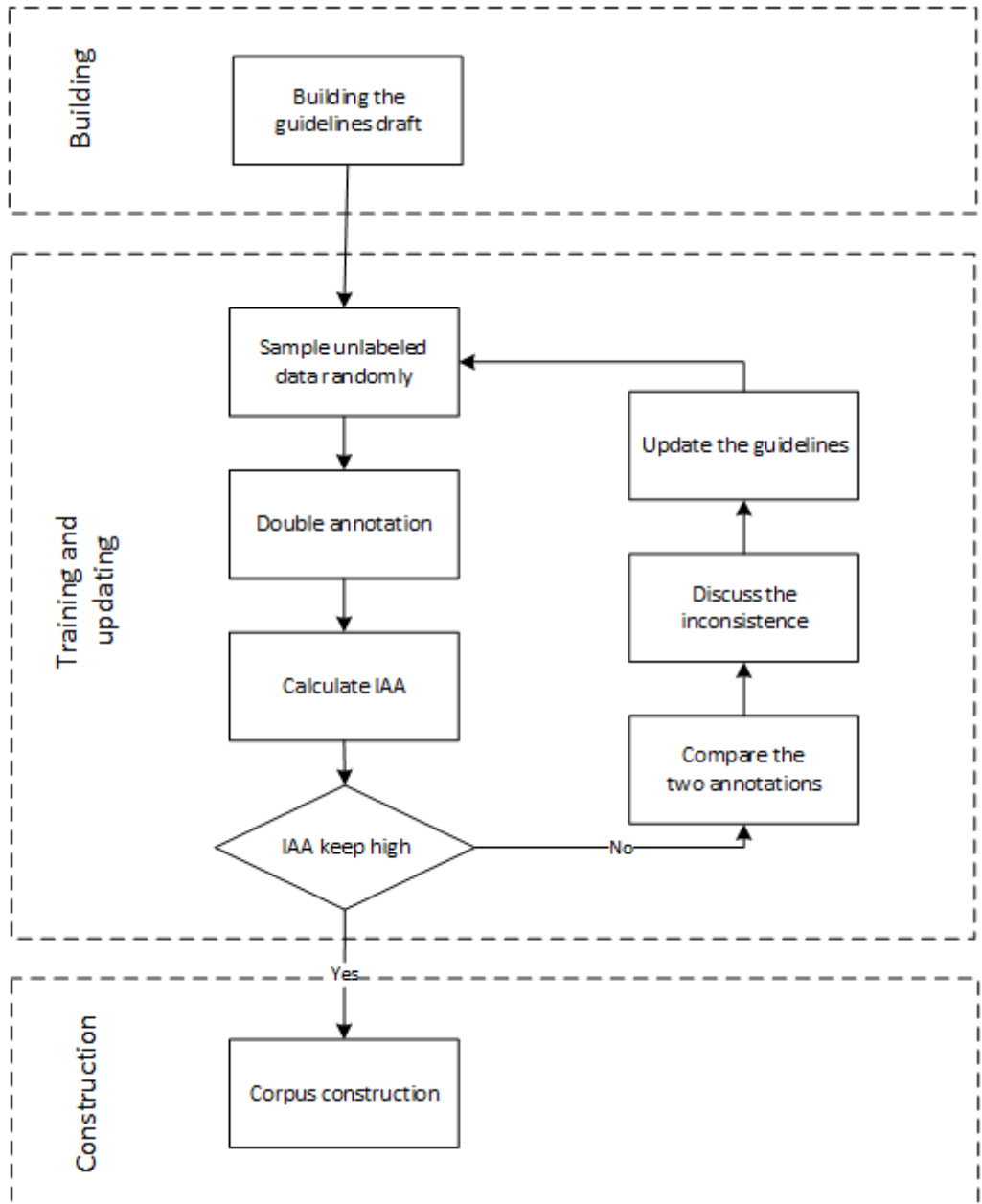

**Figure 2.** The flowchart for CVD risk factor annotation method

***Drafting the guidelines***

Based on the annotation guidelines discussed above, the linguists created a preliminary draft of the annotation guidelines that included all 12 CVD risk factors, indicators, temporal attributes, and assertions along with their definitions, as well as some positive annotations (expressions which should be marked) and negative annotations (expressions which should not be marked). Some sample annotation attempts were conducted under this draft using an annotation tool developed specifically for this task. Figure 3 shows a sample annotation.

**Figure 3.** A sample annotation for CVD risk factors

Using the sample annotations, errors and inappropriate rules in the preliminary draft were corrected, and additional positive and negative examples were added to the draft. This process continued until no further modifications were needed; at that point, the specifications were considered to be suitable for the next workflow step.

***Training the annotators and updating the guidelines***

For domain annotation, annotators with specific knowledge backgrounds are desirable. Consequently, two Masters students in medicine were employed and trained as annotators. The training process follows an iterative method, each repetition can be summed up as follows:

Phase 1: A set of discharge summaries and progress notes for 15 randomly selected patients were provided to both annotators for labeling.

Phase 2: After completing the annotation, the IAA of the two annotated corpus was calculated to evaluate the degree to which the annotators were in agreement. For the IAA calculation, one annotated database is used as the gold standard, and the other is compared to the standard to compute the precision, recall, and $F_1$-measure. Here, standard precision, recall, and $F_1$-measure equations were adopted; their calculations are as follows:

$$precision = \frac{Agreement(A1, A2)}{Annotation(A2)}, \qquad (1)$$

$$recall = \frac{Agreement(A1, A2)}{Annotation(A1)}, \qquad (2)$$

$$F_1 = \frac{2 \times precision \times recall}{precision + recall}. \qquad (3)$$

Here, we regarded the annotations of annotator $A1$ as gold standard and evaluated the quality of annotator $A2$'s annotations. The $Agreement(A1, A2)$ refers to the same annotations made by the two annotators. More calculation details can be found in Hripcsak and Rothschild [41].

Phase 3: The two annotations were compared and any uncertainties were discussed by both the linguists and the annotators. A voting method was used to obtain a final agreement.

Phase 4: The annotation guidelines were updated. In particular, errors found during phase 3 were added to the positive or negative examples and, when necessary, the guidelines were modified.

This procedure was iteratively conducted until the IAA calculated in Phase 2 achieved a continuously high value. In total, five repetitions were carried out; the resulting IAA values are listed in Table 2.

Table 2 IAA values achieved during the iterative training process

|  | Iteration 1 | Iteration 2 | Iteration 3 | Iteration 4 | Iteration 5 |
|---|---|---|---|---|---|
| Precision | 0.810 | 0.977 | 0.967 | 0.986 | 0.988 |
| Recall | 0.815 | 0.977 | 0.962 | 0.986 | 0.988 |
| $F_1$-measure | 0.812 | 0.977 | 0.964 | 0.986 | 0.988 |

As Table 2 shows, the iterations obtained very high IAAs. All the $F_1$-measures values were above 0.964 except for Iteration 1, in which the low score was probably caused by the initial unfamiliarity of annotators with the annotation guidelines and tools. Subsequent iterations obtained surprisingly high scores, indicating that the annotators and guidelines were truly ready to perform the corpus annotation.

*Corpus construction*

The annotators were asked to capture annotations from CEMRs for the 600 patients using the updated annotation guidelines. Moreover, to create a high quality annotated corpus, three measures were taken. One was that, the annotators could press a button on the annotation tool to indicate that they were unsure of the accuracy of a current annotation. Those uncertainties could be collected and discussed later. Another measure involved the use of overlapped documents (discharge summaries and progress notes of 25 patients), which were distributed to both annotators. These twice-annotated records were used to calculate the IAA and to monitor the quality of the entire annotation evaluation. The last measure was a random sampling check on the annotations (at least one third were selected) by the linguists. When problems were found, discussions were held and the guidelines were updated.

**RESULTS**

In total, in the CVD risk factor annotated corpus comprising the discharge summaries and progress notes for all the 600 patients, there are 9,678 annotations associated with the 12 CVD risk factors. Of these, the "mention" indicator type garners 63.5%, while the "drug" indicator type is rare (due to our restriction in the annotation guidelines that medication must be confirmed to be treated as a risk factor). Among the risk factors, hypertension is prominent in CEMR, with 3,729 annotations. Age and gender annotations occur at the same rate as Bp because they are basic patient attributes that are routinely recorded before diagnosis. The distributions of the four assertions (present, absent, possible, and not associated with patient) were 69.7, 19.2, 11.0, and 0.1 percent, respectively. The "present" assertion type occur most often because positive descriptions may have more significance when creating the medical records.

**Annotation quality and analysis**

Reasonably, the IAA values for the final corpus should be as high as the IAA values obtained during the training process due to the work performed before and during the formal annotation to guarantee sufficient quality. The final IAA calculations resulted in 0.971 for precision, 0.965 for recall and 0.968 for the $F_1$-measure; these values demonstrate the high quality of the corpus.

Table 3 shows the distribution of risk factors, indicators, temporal attributes and assertions. Each row in the table shows the distribution of a single indicator over the entire corpus in different time and assertion partitions.

--Table 3 about here--

For hypertension, "high Bp" is the most common indicator and "mention" is second. Approximately 2 "mention" and 3 "high Bp" indicators appear in each patient's records. The "mention" annotations are mostly continuing and usually fixed over short durations, while "high Bp" annotations tend to occur during DHSs because Bp measurements are taken during physical examinations. Meanwhile, 78.7 percent of the "regulate Bp" annotations occur during DHSs because controlling Bp is a standard treatment for hypertensive individuals.

Compared with "high Bp", annotations identifying "high blood glucose" are far less common because of the complicated testing technique for blood glucose. In Table 3, the "mention" annotations comprise 86.7% of all the diabetes annotations for this risk factor, and 64.1% of these "mention" annotations are "continuing, absent." From earlier discoveries in the records, we knew that "否认高血压、糖尿病病史" (denying a history of hypertension and diabetes) occurs frequently.

The spotlight indicator of dyslipidemia annotations is "regulate blood lipids", because "调节血脂" (to regulate blood lipids) is a representative narrative in the assessment and plan section of dyslipidemia records. Moreover, for the same reason, the timing of this indicator is clustered around "during DHS, present".

When considering CKD, OSAS and FHCVD, the only indicator "mention" occurs infrequently. It occurs the most with CKD but still fewer than 26 times. Meanwhile, there is no "absent" assertion for any of these annotations.

Atherosclerosis for "稳定冠脉粥样斑块" (stabilizing coronary atheromatous plaque)

has a relatively high number of mentions in our corpus and occurs repeatedly in the assessment and planning sections of CEMRs.

The number of smoking annotations is relatively high (508 annotations among the records of 600 patients). "Mention" and "smoking amount" account for almost all the occurrences. The 380 "mention" and 119 "smoking amount" annotations include the "continuing" assertion because tobacco use is generally a habit, and quitting rarely occurs over a short period.

There are numerous references to alcohol in the narratives of CEMRs, such as "否认吸烟、饮酒史" (denying the history of smoking and drinking) and "间断少量饮酒史" (a history of intermittent small amounts of alcohol). However, these were not tagged as alcohol abuse because the patient's intake is none or a slight. In contrast, serious usage has only 76 "mention" and 19 "drinking amount" annotations in our corpus and nearly all those are continuing.

Age and gender are rich basic information in CEMRs. As with actual discoveries in the narratives of CEMRs, most "mention" annotations for age and gender occur in the hospitalization information section of discharge summaries and in the complaint, case characteristics and diagnosis basis sections of progress notes. Occasionally, "age group" occurs in the case characteristics and diagnosis basis sections.

**DISCUSSION**

We developed a corpus of CVD risk factor annotations including indicators, temporal attributes and assertions. Linguists and clinicians cooperated throughout the entire corpus construction process—from drafting annotation guidelines to discussing disagreements. The final IAA values achieved for this corpus reflects its high level.

In our corpus, a test value was annotated whether a test outcome was above or below a standard threshold. This was intentional and was designed to build a complete record of risk factor variation over the long term. For hypertension, we annotated all the Bp values regardless of whether they were above the standard 140/90 mmHg. Based on these annotations, when supervising an individual's long-term health condition, the trained IE system can extract all the Bp conditions with no omissions and can then provide a visualization of the Bp variations over time. An appropriate warning or intervention treatment could then be applied at critical variation points.

Annotations of therapeutic methods such as blood glucose regulation and hypoglycemic drug administration supports optional treatment recommendations. The trained IE system, after extracting the information from large numbers of CEMRs, can provide clinicians with referable treatments when it finds a similar condition in a current patient. Along a time dimension, these extractions can provide a clear assessment of treatment effects, for example, when a need to regulate glucose was recognized but glucose treatments were later stopped could show that the regulating treatment had a positive effect. This is significant because it allows clinicians to treat similar patients and decide which treatment would be better.

The annotations of O2, hypertension, diabetes, dyslipidemia, CKD, and Atherosis can play a role in managing these chronic diseases. An IE system can extract an individual's characteristics, such as examination values and medications. Then, a long term monitoring platform can monitor variations in these characteristics, provide feedback on treatments effects, and predict disease tendencies. Globally, nearly one billion people have high blood pressure; in 2008, diabetes was responsible for 1.3 billion deaths; high cholesterol was estimated to have caused 2.6 million deaths; and at least 2.8 million people die each year as a result of being overweight or obese [5]. Consequently, a monitoring platform that aids in managing these chronic diseases can significantly reduce the suffering they cause to many patients and, thus, reduce CVD.

**CONCLUSION**

This paper describes the construction of an annotated corpus of CVD risk factors in CEMRs. To the best of our knowledge, this is the first Chinese corpus that concerns risk factors for CVD. We engaged both clinicians and annotators to draft guidelines and annotate the medical records. We proposed an annotation method that results in high quality annotations; the presented IAA values indicate the high quality of the resulting corpus. These document-level risk factor annotations, along with the included temporal attributes and assertions, can be utilized in future studies of risk factor progression and their relationships with CVD over time. This corpus can play a significant role in developing a future IE system that can extract CVD risk factors from CEMRs to build a clear picture of individuals' CVD risk factors and conditions, and it makes developing a monitoring platform to supervise the progression of risk factors and CVD possible. The related annotation resources are publicly available at https://github.com/WILAB-HIT/RiskFactor.

**LIST OF ABBREVIATIONS**

CVD: cardiovascular disease; CEMRs: Chinese electronic medical records; NLP: natural language processing; EMR: electronic medical records; IE: information extraction; i2b2: Integrating Biology and the Bedside; POS: part-of-speech; IAA: inter-annotated agreement;

ICU: Intensive Care Unit; OCR: optical character recognition; XML: Extensible Markup Language; Bp: blood pressure; O2: Overweight/Obesity; CKD: Chronic Kidney Disease; OSAS: Obstructive Sleep Apnea Syndrome; A2: Alcohol Abuse; FHCVD: Family History of CVD; RBG: Random Blood Glucose; TG: triglyceride; DHS: duration of hospital stay

# DECLARATIONS

**Ethics approval**

The ethical approval is available at https://github.com/WILAB-HIT/RiskFactor/tree/master/statement_ of_the_ethics.

**Consent for publication**

Not applicable.

**Availability of data and materials**

The datasets used and analysed during the current study available from the corresponding author on reasonable request.

**Competing interests**

None.

**Funding**

None.

**Authors' contributions**

JS, BH and JY designed the tasks and conducted the entire project. JS, BH and JJ developed the annotation guidelines and collaborated in constructing the corpus. JS and BH distributed the data, collected the annotations, analyzed the corpus quality and communicated with the doctors. All the authors contributed to the final manuscript.

**Acknowledgements**

We would like to thank the medical records department of the 2nd Affiliated Hospital of Harbin Medical University for providing the electronic medical records. We would also like to thank the clinicians, Qiubin YU and Yongjie Zhao, and the annotators, Hao Wu and Na Feng, for their excellent work.

**Authors' information**

Jia Su, PhD candidate, school of computer science and technology, Harbin institute of technology, mainly study the information extraction and machine learning based on Chinese electronic medical records.

Bin He, PhD candidate, school of computer science and technology, Harbin institute of technology, mainly study the named entities recognition and machine learning based on Chinese electronic medical records.

Yi Guan, doctor and professor, school of computer science and technology, Harbin institute of technology, mainly study the natural language processing, medical health informatics and network mining.

Jingchi Jiang, PhD candidate, school of computer science and technology, Harbin institute of technology, mainly study the knowledge representation and probabilistic graphical model.

Jinfeng Yang, doctor, School of Software, Harbin University of Science and Technology, mainly study the natural language processing and machine learning.

**TABLES**

**Table 3** Distribution of CVD risk factors, indicators, their occurrence times, and assertions

| Risk factors | Indicators | Before DHS | | | | | During DHS | | | | | After DHS | | | | | Continuing | | | | | Total |
|---|---|---|---|---|---|---|---|---|---|---|---|---|---|---|---|---|---|---|---|---|---|---|
| | | P | A | Pb | N | Total | P | A | Pb | N | Total | P | A | Pb | N | Total | P | A | Pb | N | Total | |
| O2 | Mention | 0 | 0 | 0 | 0 | 0 | 0 | 0 | 0 | 0 | 0 | 0 | 0 | 0 | 0 | 0 | 18 | 0 | 0 | 0 | 18 | 18 |
| Hypertension | Mention | 0 | 1 | 0 | 0 | 1 | 0 | 0 | 1 | 0 | 1 | 0 | 0 | 0 | 0 | 0 | 526 | 401 | 471 | 0 | 1398 | 1400 |
| | High Bp | 304 | 0 | 0 | 0 | 304 | 1647 | 0 | 0 | 0 | 1647 | 0 | 0 | 0 | 0 | 0 | 2 | 1 | 1 | 0 | 4 | 1955 |
| | Regulate Bp | 56 | 0 | 0 | 0 | 56 | 244 | 0 | 0 | 0 | 244 | 10 | 0 | 0 | 0 | 10 | 0 | 0 | 0 | 0 | 0 | 310 |
| | Blood pressure drug | 43 | 1 | 0 | 0 | 44 | 17 | 0 | 0 | 0 | 17 | 0 | 0 | 0 | 0 | 0 | 0 | 3 | 0 | 0 | 3 | 64 |
| Diabetes | Mention | 2 | 0 | 0 | 0 | 2 | 1 | 0 | 0 | 0 | 1 | 0 | 0 | 0 | 0 | 0 | 172 | 558 | 138 | 3 | 871 | 874 |
| | High blood glucose | 19 | 0 | 0 | 0 | 19 | 20 | 0 | 0 | 0 | 20 | 0 | 0 | 0 | 0 | 0 | 4 | 0 | 0 | 0 | 4 | 43 |
| | Regulate blood glucose | 8 | 0 | 0 | 0 | 8 | 28 | 0 | 0 | 0 | 28 | 6 | 0 | 0 | 0 | 6 | 0 | 0 | 0 | 0 | 0 | 42 |
| | Hypoglycemic drug | 31 | 0 | 0 | 0 | 31 | 8 | 0 | 0 | 0 | 8 | 7 | 0 | 0 | 0 | 7 | 2 | 0 | 0 | 0 | 2 | 48 |
| Dyslipidemia | Mention | 0 | 0 | 0 | 0 | 0 | 0 | 0 | 0 | 0 | 0 | 0 | 0 | 0 | 0 | 0 | 45 | 4 | 24 | 0 | 73 | 73 |
| | High blood lipids | 0 | 0 | 0 | 0 | 0 | 4 | 0 | 0 | 0 | 4 | 0 | 0 | 0 | 0 | 0 | 2 | 0 | 0 | 0 | 2 | 6 |
| | Regulate blood lipids | 2 | 0 | 0 | 0 | 2 | 249 | 0 | 0 | 0 | 249 | 5 | 0 | 0 | 0 | 5 | 0 | 0 | 0 | 0 | 0 | 256 |
| | Lip-lowering drug | 2 | 0 | 0 | 0 | 2 | 34 | 0 | 0 | 0 | 34 | 1 | 0 | 0 | 0 | 1 | 0 | 0 | 0 | 0 | 0 | 37 |
| CKD | Mention | 0 | 0 | 0 | 0 | 0 | 0 | 0 | 0 | 0 | 0 | 0 | 0 | 0 | 0 | 0 | 9 | 0 | 17 | 0 | 26 | 26 |
| Atherosis | Mention | 3 | 0 | 0 | 0 | 3 | 4 | 0 | 0 | 0 | 4 | 0 | 0 | 0 | 0 | 0 | 136 | 0 | 1 | 0 | 137 | 144 |
| OSAS | Mention | 0 | 0 | 0 | 0 | 0 | 0 | 0 | 0 | 0 | 0 | 0 | 0 | 0 | 0 | 0 | 1 | 0 | 0 | 0 | 1 | 1 |
| Smoking | Mention | 0 | 0 | 0 | 0 | 0 | 0 | 0 | 0 | 0 | 0 | 0 | 0 | 0 | 0 | 0 | 231 | 149 | 0 | 0 | 380 | 380 |
| | Smoking cessation | 5 | 2 | 0 | 0 | 7 | 0 | 0 | 0 | 0 | 0 | 0 | 0 | 0 | 0 | 0 | 1 | 0 | 0 | 0 | 1 | 8 |
| | Smoking amount | 0 | 1 | 0 | 0 | 1 | 0 | 0 | 0 | 0 | 0 | 0 | 0 | 0 | 0 | 0 | 119 | 0 | 0 | 0 | 119 | 120 |
| A2 | Mention | 9 | 0 | 0 | 0 | 9 | 0 | 0 | 0 | 0 | 0 | 0 | 0 | 0 | 0 | 0 | 52 | 15 | 0 | 0 | 67 | 76 |
| | Drinking amount | 0 | 0 | 0 | 0 | 0 | 0 | 0 | 0 | 0 | 0 | 0 | 0 | 0 | 0 | 0 | 19 | 0 | 0 | 0 | 19 | 19 |
| FHCVD | Mention | 0 | 0 | 0 | 0 | 0 | 0 | 0 | 0 | 0 | 0 | 0 | 0 | 0 | 0 | 0 | 10 | 0 | 0 | 0 | 10 | 10 |
| Age | Mention | - | - | - | - | - | - | - | - | - | - | - | - | - | - | - | - | - | - | - | - | 1233 |
| | Age group | - | - | - | - | - | - | - | - | - | - | - | - | - | - | - | - | - | - | - | - | 626 |
| Gender | Mention | - | - | - | - | - | - | - | - | - | - | - | - | - | - | - | - | - | - | - | - | 1909 |

P: Present, A: Absent, Pb: Possible, N: Not associated with the patient. "-" denotes not considered.